\newif\ifarxiv

\arxivtrue

\ifarxiv
	\documentclass[final, twocolumn]{elsarticle}
\else
  \documentclass[preprint,12pt]{elsarticle}
  \usepackage{lineno}
  \modulolinenumbers[5]  
\fi

\usepackage{amssymb}
\usepackage{graphicx}
\usepackage{longtable}
\usepackage{multirow}
\usepackage{amsmath}
\usepackage{url}
\usepackage{hyperref}
\usepackage{adjustbox}	
\hypersetup{
     colorlinks   = true,
     citecolor    = blue
}
\usepackage[usenames,dvipsnames,svgnames,table]{xcolor}

\ifarxiv
\journal{Neurocomputing}
\else
\journal{Neurocomputing}
\fi

\begin{document}


\begin{frontmatter}

\title{MirBot: A collaborative object recognition system for smartphones using convolutional neural networks}

\author{Antonio Pertusa\corref{cor1}}
\ead{pertusa@dlsi.ua.es}
\author{Antonio-Javier Gallego}
\ead{jgallego@dlsi.ua.es}
\author{Marisa Bernabeu}
\ead{mbernabeu@dlsi.ua.es}
\address{Dpto. Lenguajes y Sistemas Inform\'{a}ticos, Universidad de Alicante, E-03690, San Vicente del Raspeig, Alicante, Spain}
\cortext[cor1]{Corresponding author.}



\begin{abstract}
MirBot is a collaborative application for smartphones that allows users to perform object recognition. This app can be used to take a photograph of an object, select the region of interest and obtain the most likely class (dog, chair, etc.) by means of similarity search using features extracted from a convolutional neural network (CNN). The answers provided by the system can be validated by the user so as to improve the results for future queries. All the images are stored together with a series of metadata, thus enabling a multimodal incremental dataset labeled with synset identifiers from the WordNet ontology. This dataset grows continuously thanks to the users' feedback, and is publicly available for research. This work details the MirBot object recognition system, analyzes the statistics gathered after more than four years of usage, describes the image classification methodology, and performs an exhaustive evaluation using handcrafted features, neural codes, different transfer learning techniques, PCA compression and metadata, which can be used to improve the image classifier results. 
The app is freely available at the Apple and Google Play stores. 
\end{abstract}



\end{frontmatter}

\ifarxiv
\else
\linenumbers
\fi

\section{Introduction}

Object recognition is a highly active topic in computer vision and can be particularly useful for mobile devices \citep{Bock2010IImage:IPhone, mobilenets} as regards retrieving information about objects on the fly. Visually impaired persons can also benefit from these systems \citep{Matusiak2013ObjectUsers}. 

\begin{figure*}
\centering
\includegraphics[scale=0.55]{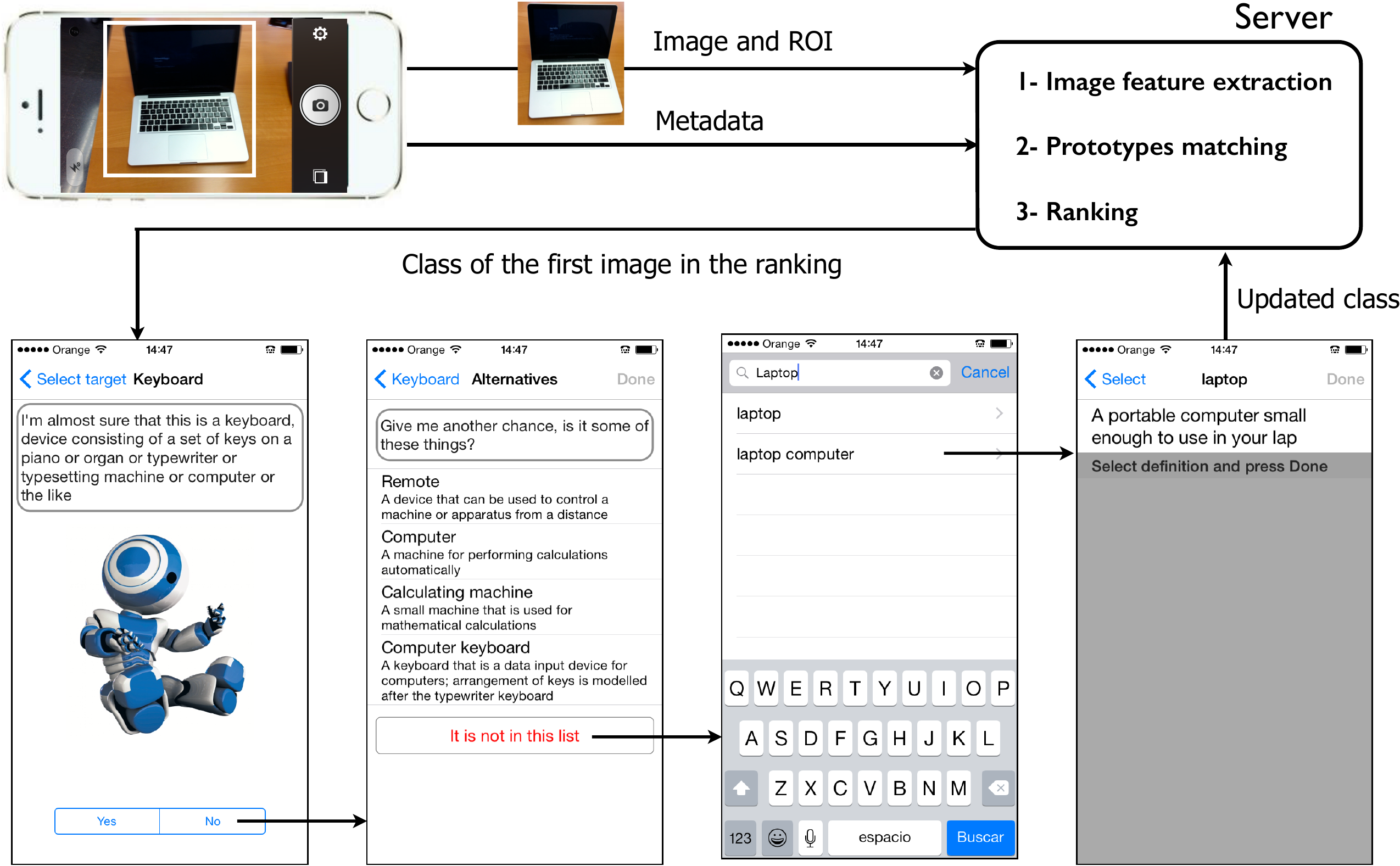}
\caption{\label{Fig1} Architecture of the iOS app. This example corresponds to the longest user interaction sequence. }
\end{figure*}

MirBot is a smartphone app that users can train to recognize any object. The objects are categorized according to lemmas (such as chair, dog, laptop, etc.) selected from the WordNet lexical database \citep{Fellbaum1998WordNet:Database}. A user can employ MirBot to take a photograph and select a rectangular region of interest (ROI) in which the target object is located. The image, the ROI coordinates and a series of associated metadata are then sent to a server, which performs a similarity search using k-Nearest Neighbors (kNN) and returns the class of the most likely image, as shown in Figure \ref{Fig1}. The app users can validate the system response in order to improve the classification results for future queries, and this feedback allows the database to grow continuously with new labeled images. 

MirBot is designed as a pedagogical game in which a simulated robot can be trained in order to involve users in an automatic learning process. As pointed out in \cite{Barrington2012Game-poweredLearning.,vonAhn2004LabelingGame}, developing machine learning tasks through games has proven to be a successful approach to make users enjoy labeling data. From the users' perspective, the main distinctive feature of MirBot with regard to other apps is that it allows them to train a personal image search system, thus making the dataset dynamic and user driven.

This work extends the contents of the MirBot system for object retrieval introduced in \cite{MirBot:System}, which initially used handcrafted visual descriptors. The main contributions of this paper with regard to the initial work are a detailed description of the user interaction process, the statistics related to the database gathered after four years of usage, a new classification methodology based on CNN features (neural codes) obtained from pre-trained and fine-tuned models, the study of PCA compression on different neural networks, the inclusion of metadata to complement the neural codes and the evaluation results and conclusions. 

When a user submits a photograph, visual descriptors are extracted and compared to the existing prototypes in the dataset in order to predict the class of the object. In the initial MirBot version \cite{MirBot:System}, both local features and color histograms were extracted and combined to obtain the most likely class. In this work, several convolutional neural network (CNN) features have been added to these descriptors for use in evaluation. The gap between the results obtained using handcrafted descriptors and features extracted from convolutional networks led the traditional image descriptors in MirBot to be replaced with neural codes in June 2015.

As pointed out in \cite{Torralba200880Recognition.}, finding images within large collections is an important topic for the computer vision community. Recent progress in object recognition has been built upon efforts to create large-scale, real-world image datasets \citep{Khosla2012UndoingBias, Deng2009ImageNet:Database} that are crucial for developing robust image retrieval algorithms, in addition to considering the large amount of data required in recent deep neural networks \citep{LeCun2015DeepLearning}. 

One of the main contributions of MirBot is a dataset with a similar structure to that of ImageNet \citep{Deng2009ImageNet:Database}, with the exception that all the images are gathered with smartphone cameras and stored with their associated metadata and with regions of interest. In October 2016 we had $25,292$ validated images distributed in $1,808$ classes. Although the MirBot dataset still cannot be considered a very large collection, it is incremental and grows continuously thanks to its users' feedback. One important difference with regard to other datasets such as \cite{Torralba200880Recognition.} and \cite{Deng2009ImageNet:Database} is that, rather than employing images downloaded from the Internet, users take pictures specifically for object recognition. This signifies that MirBot images are focused on the target objects and gathered with minimum occlusions and plain backgrounds. Our team reviews the new images on a weekly basis in order to avoid inappropriate, unfocused or wrongly labeled samples, thus ensuring good quality data. 

The MirBot dataset also includes a series of metadata that could be used to constrain the search space. These metadata, which are detailed in \cite{MirBot:System}, are extracted from the smartphone sensors (angle with regard to the horizontal, gyroscope, flash, GPS, etc.), and have reverse geocoding information (type of place, country, closest points of interest, etc.) and EXIF camera data (aperture, brightness, ISO, etc.). All the images are stored with their associated metadata. We have evaluated the performance using these metadata in order to complement the image information.

This work begins by describing the user interaction interface in Section \ref{interaction}, and the methodology (Section \ref{methodology}) used for similarity search on the server side. The evaluation results are described and discussed in Section \ref{evaluation}, followed by the conclusions in Section \ref{conclusions}.

\section{User interaction}
\label{interaction}

\begin{figure*}
\center
\includegraphics[width=\textwidth]{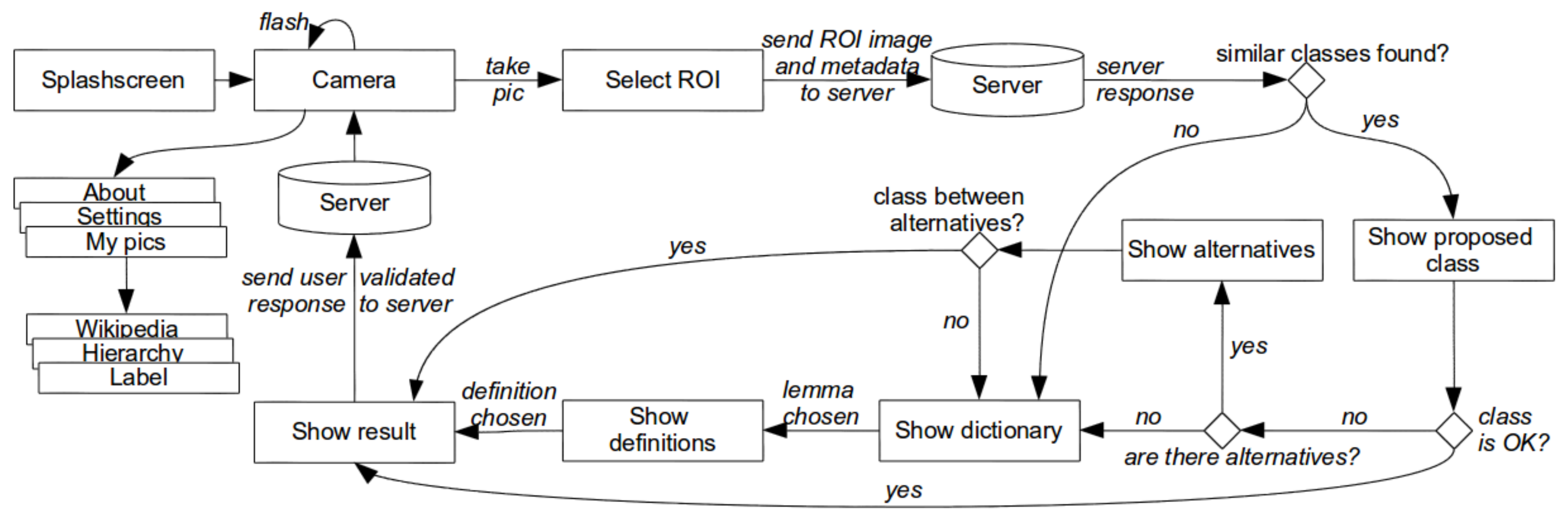}
\caption{\label{Fig2} Diagram of the MirBot interface. The path from \texttt{Server} (after ROI selection) to \texttt{Server} corresponds to the user validation process. }
\end{figure*}

As stated in \cite{Lew2006Content-basedChallenges}, beyond the one-shot queries in the early similarity-based search systems, the next generation of systems attempts to integrate continuous feedback from the user in order to learn more about the query. MirBot is designed as an interactive game and the objective of its interface is to minimize the number of user interactions. 

In order to avoid the dispersion that can be caused when using free object identifiers, users can only assign class names selected from the WordNet ontology. The main advantage of WordNet is its semantic structure, which prevents ambiguities in the labels. WordNet \textit{synsets} (synonym sets) are unique identifiers for meaningful semantic concepts, and each of them is linked to a definition, although they can be related to different lemmas. Similarly to ImageNet, we use the WordNet synsets as class identifiers.

In MirBot, only portrayable objects can be chosen from the WordNet hierarchy, including the following root categories: animals (lexicographer identifier: $5$), food/drinks ($13$), plants ($20$), and objects, which include both WordNet natural objects ($6$) and artifacts ($17$). WordNet considers that an artifact is man-made whereas a natural object is not, but they have been merged in MirBot in order to simplify this concept for its users. 

Figure \ref{Fig2} shows the block diagram of the user interface. Before sending the query image to the server, users can find settings in the app that allow to choose whether they wish to perform the classification by considering only their own images, so as to constrain the search space, or by using the whole dataset.  

Once the query has been submitted, the classes of the $K$ most similar images are retrieved using the methodology described in Section \ref{methodology}, and the class of the first image in this ranking is given to the user for validation. If no images in the dataset are found to be similar, then an unknown object message is displayed and the user is asked to select a class from the WordNet dictionary. 

In the case of the user confirming the server response, the image is labeled with that class and stored in the dataset. Otherwise, an alternative list containing the classes of the top-$K$ images in the ranking is shown. If none of them correspond to the actual class, the user can select a lemma (e.g., \texttt{key}) from the WordNet nouns, and then a definition related to this lemma (e.g.,  \texttt{Metal device shaped...}), which corresponds to the synset (class identifier). 

The messages and the robot images in \texttt{Show proposed class} (see Figure \ref{Fig2}) are dependent on the distance $\min_{\forall{b}} {d(a,b)}$ between the query image $a$ and the most likely prototype $b$, in order to enhance the user’s experience. 

Two constant thresholds are used for this purpose: $\lambda$, which indicates a certain answer (a very small distance from the query image to its closest prototype), and  $\theta>\lambda$, which indicates that the object is unknown (a very large distance to its closest prototype). 

If $\min_{\forall{b}} {d(a,b)} < \lambda$, then the message begins with \textit{I'm pretty sure that this is...}. If $ \lambda \le \min_{\forall{b}}{d(a,b)} \le \theta$, then the range $[\lambda, \theta]$ is discretized into $10$ levels that trigger messages such as \textit{I'm almost sure that...}, \textit{I have many doubts, but ...}, etc. Finally, if $\min_{\forall{b}} {d(a,b)} > \theta$, then the user receives an answer indicating the object is unknown, as there are no prototypes in the database  that are sufficiently close.

The messages and robot images in \texttt{Show result} similarly depend on the validated answer. If the proposed class is not correct, the closest common level in the WordNet hierarchy determines the system response (\textit{I was close}, \textit{I was totally wrong}, etc.). 

Besides setting the WordNet class, users can also manage their images in \texttt{My pics} and label them using any text (e.g. \textit{My dog Toby}).  If the closest image in the ranking contains a label, it is also shown to the user in \texttt{Show proposed class}. Users can also learn more about their objects with Wikipedia and consult the WordNet hierarchy of a synset.

In the Android version, some gaming techniques were implemented so as to involve more users. These additions include an IQ measure for the robots, a ranking of the best players and achievement badges. The next version of the iOS app will also include these features.

\section{Methodology}
\label{methodology}

After a photograph has been taken and sent to the server, object classification is performed in order to return an answer to the user. This section describes the image features and the similarity search techniques used to retrieve the most likely class. At present, only the image contents within the ROI are used to extract the visual descriptors in order to perform the classification.

\subsection{Visual descriptors}
\label{features}

Visual descriptors are extracted from the image on the server side. It is now also feasible to integrate feature extraction into a smartphone app \cite{mobilenets}. However, performing this on the server increases the flexibility of the system, as the descriptors and classifiers can easily be updated without having to modify the source code on the client side. Moreover, the incremental nature of the proposed method (new classes are continuously added) requires that the information must be centralized. 

Traditional local invariant descriptors \citep{Mikolajczyk2004ScaleDetectors}, such as SIFT   \cite{Lowe2004DistinctiveKeypoints} and SURF \cite{Bay2008Speeded-UpSURF}, and global descriptors such as color histograms \citep{vandeSande2010EvaluatingRecognition, Jeong2001Histogram-BasedRetrieval} have been extensively used for object recognition in literature. Some methods also combined different features, such as color with local descriptors \citep{Fernando2012DiscriminativeClassification}, or shape with color and texture \citep{Banerji2013NewClassification}. 

However, since 2013 most systems have relied on deep learning, as convolutional neural networks have dramatically improved the state of the art of visual object recognition. As pointed out in \cite{LeCun2015DeepLearning}, the ability of these techniques to learn the representations directly from the raw data  have revolutionized many classification tasks related to image, video, speech and audio.

The visual features used in the initial MirBot version \cite{MirBot:System} were TopSURF local descriptors \citep{Thomee2010TOP-SURF:Toolkit} and global YCbCr color histograms weighted with a two-dimensional Gaussian function. These features were replaced with neural codes in July 2015, as they clearly outperformed handcrafted descriptors. This section describes the different features evaluated using the MirBot dataset.

\subsubsection{Local handcrafted descriptors}

In \cite{MirBot:System} we initially used the TopSURF \citep{Thomee2010TOP-SURF:Toolkit} toolkit to obtain a histogram of local descriptors for each image. This method calculates the Speed-Up Robust Features (SURF \citep{Bay2008Speeded-UpSURF} interest points), and clusters them \citep{Philbin2007ObjectMatching} into a bag of features so as to yield a dictionary of visual words. Those visual words that do not occur very frequently are emphasized using a tf-idf \citep{Salton1983IntroductionRetrieval.} weighting technique. Finally, the descriptor consists of a tf-idf histogram obtained by selecting those visual words with the highest score (the top $T$ words). 

Generic dictionaries\footnote{\texttt{http://press.liacs.nl/researchdownloads/topsurf/}} of different sizes $D$ were evaluated in order to calculate the tf-idf histograms, as described in Section \ref{evaluation}. A non-generic dictionary trained with the MirBot data was also used for comparison purposes.

\subsubsection{Global handcrafted descriptors}

Standard SURF features do not consider color \cite{Jeong2001Histogram-BasedRetrieval}, which can be relevant when attempting to identify certain classes. The TopSURF descriptors in the initial version of MirBot \cite{MirBot:System} were, therefore, complemented with color histograms in order to improve accuracy. 

Experiments with which to evaluate different color spaces (RGB, HSV, CIE-LUV and YCrCb) were carried out, and the best results were obtained using YCbCr, which has proven to be relatively robust to changes in lighting. The color value of each pixel was additionally weighted using a two-dimensional Gaussian function to obtain a weighted color histogram, as detailed in \cite{MirBot:System}. The goal of this weighting is to give less relevance to the colors that appear on the edges of the image and more weight to those on the middle, which is where the objects to be recognized are likely located, given their ROI.

\subsubsection{Neural codes using pretrained models}
\label{sec:ncpm}

Transfer learning is a common technique consisting of training a network using a given dataset and repurposing (or transferring) the features learned to a different dataset or network \cite{Yosinski2014How}. This process usually obtains remarkable results when the features are general (suitable for both base and target tasks).

We have evaluated several CNN models in order to obtain the neural codes: 

\begin{itemize}
\item AlexNet \cite{Krizhevsky2012ImageNetNetworks} is a CNN with 8 layers. Since it outperformed previous methods by a large margin in the ImageNet ILSVRC12 \citep{Russakovsky2014ImageNetChallenge} challenge (which contains about 1.3 million images distributed in 1,000 object categories), most image recognition systems rely on CNNs. 

\item GoogLeNet \cite{Szegedy2014GoingConvolutions} is a well-known CNN which achieved the best results in the ImageNet ILSVRC14 challenge. This network contains 22 layers but uses 12 times fewer parameters than AlexNet, and is based on a series of chained Inception modules. 

\item VGG-16 and VGG-19 \cite{Simonyan14}. VGG-16 has 13 convolutional and 3 fully-connected layers, whereas VGG-19 is composed of 16 convolutional and 3 fully-connected layers. Both topologies use Dropout and Max-pooling techniques and ReLU activation functions. 

\item Inception21k is an Inception v2 network based on a GoogleNet with Batch Normalization \cite{Ioffe15:ArXiv}. We call this network Inception21k because it was trained with the full ImageNet dataset (14,197,087 images in 21,841 classes), unlike AlexNet and GoogleNet, which were trained with the 1,000 classes from the ILSVRC challenge. 

\item Inception v3 \cite{inceptionv3}. This architecture has 6 convolutional layers followed by 3 inception modules and a last fully connected layer. It has fewer parameters than other similar models thanks to the Inception modules whose design is based on two main ideas: The approximation of a sparse structure with spatially repeated dense components and the use of dimensionality reduction to maintain the computational complexity within bounds.

\item REsNet \cite{resnet}. The deep REsidual learning Network learns residual functions with reference to the layer inputs rather than learning unreferenced functions. This technique enables the use of a large number of layers. We have used the 50-layer version for our experiments.  

\item Xception \cite{xception}. This model has 36 convolutional layers with a redesigned version of Inception modules which enable the depth-wise separable convolution operation. This architecture outperforms the Inception results using the same number of parameters. 
\end{itemize}

We have used the AlexNet and GoogLeNet implementations from Caffe \citep{Jia2014CaffeEmbedding} as a basis. These are almost identical to the original papers, with pretrained models from the ImageNet ILSVRC12 dataset. The original Inception21k network was trained by Mu Li for MXNet \citep{MXNet}, although we converted it to the Caffe framework by translating the MXNet \texttt{Batch Normalization} layers to Caffe \texttt{BatchNorm} layers with the learned mean and variance, followed by a \texttt{Scale} layer that applies the learned scale ($\gamma$) and shift ($\beta$). We have publicly released this model at \url{https://github.com/pertusa/InceptionBN-21K-for-Caffe}. In the case of the VGG, Xception, ResNet and Inception v3, we used the Keras \cite{keras} implementations with their default parameters.

The target images are forwarded through the pretrained network in order to obtain the visual descriptors, which are vectors containing the neuron output activations of the last hidden layer (excluding the output layer). For example, in the case of AlexNet, this is the Caffe layer \texttt{fc7} with $4,096$ values, in GoogLeNet it is \texttt{pool5/7x7\_s1} with a dimension of $1,024$, and in Inception21k it is the \texttt{global\_pool} layer, which also has a dimension of $1,024$. 

The main advantages of using the neural codes with kNN rather than the output layer are that the model can be used incrementally, the output classes can be different to those used for training, and we also obtain a ranking of the most similar images.

\subsection{Prototype matching}
\label{classification}

Given a sample query, the class of the most likely image among the set of prototypes is given to the user for validation. In this scenario, the techniques chosen for similarity search must be efficient and incremental owing to the real-time constraints of the proposed architecture, as users cannot wait long for a system response. 

As efficiency is crucial for our system and the database is unbalanced (there are many more samples of some classes than of others \cite{He09}), an approximate nearest neighbor search technique has been chosen to match the image descriptors from a query image with the prototypes in the dataset. 

This is done using the Spotify Annoy approximate kNN method \cite{annoy}. In the MirBot dataset, a good compromise between accuracy and performance was experimentally found by using an index of $100$ trees and setting the maximum number of nodes to $1,000$. As MirBot should operate in real time, the index cannot be rebuilt (this takes about 1 min) every time a new image is added. The descriptors that were not stored are, therefore, kept in the tree index in a database table, and the kNN is obtained by performing the search in both the Annoy indexed prototypes and in the descriptor table. The Annoy index is rebuilt weekly.

In our previous work \cite{MirBot:System}, the normalized cosine similarity was used to compare the TopSURF descriptors of two images, as in \cite{Thomee2010TOP-SURF:Toolkit}, whereas color histograms were compared using the Jensen-Shannon divergence \citep{Lin1991DivergenceEntropy}. Neural codes were compared using the Euclidean distance.

Given a query image $a$, its $K=10$ nearest neighbor images from the set of prototypes are retrieved according to their distance $d(a,b)$. The classes of the top-$K$ images in this ranking with distances $d(a,b)<\theta$  are then sent to the user for validation. As stated in the previous section, if no images in the dataset are found to be similar ($\min_{\forall{b}} {d(a,b)>\theta}$), then an unknown object message is displayed and the user is asked to select a class from the WordNet dictionary. 

In the initial MirBot system, TopSURF and color descriptors were also combined to yield a distance $d(a,b)$ between two images:

\begin{equation}
\label{eqWeight}
d(a,b)=w \cdot d_{t}(a,b) + (1-w) \cdot d_{c}(a,b)
\end{equation}\noindent where $d_t$ is the TopSURF distance, $d_c$ is the color distance, and $w$ is a parameter used to weight their contribution in the final distance $d(a,b)$.

\section{Evaluation}
\label{evaluation}

\subsection{Dataset}

The dataset is freely available on request through a web interface at \url{http://www.mirbot.com/admin}. Researchers can explore the images by means of an expandable tree view, and download them with their metadata and features.

\begin{figure*}
\centering
  \begin{tabular}{@{}cc@{}}
    \includegraphics[width=.45\textwidth]{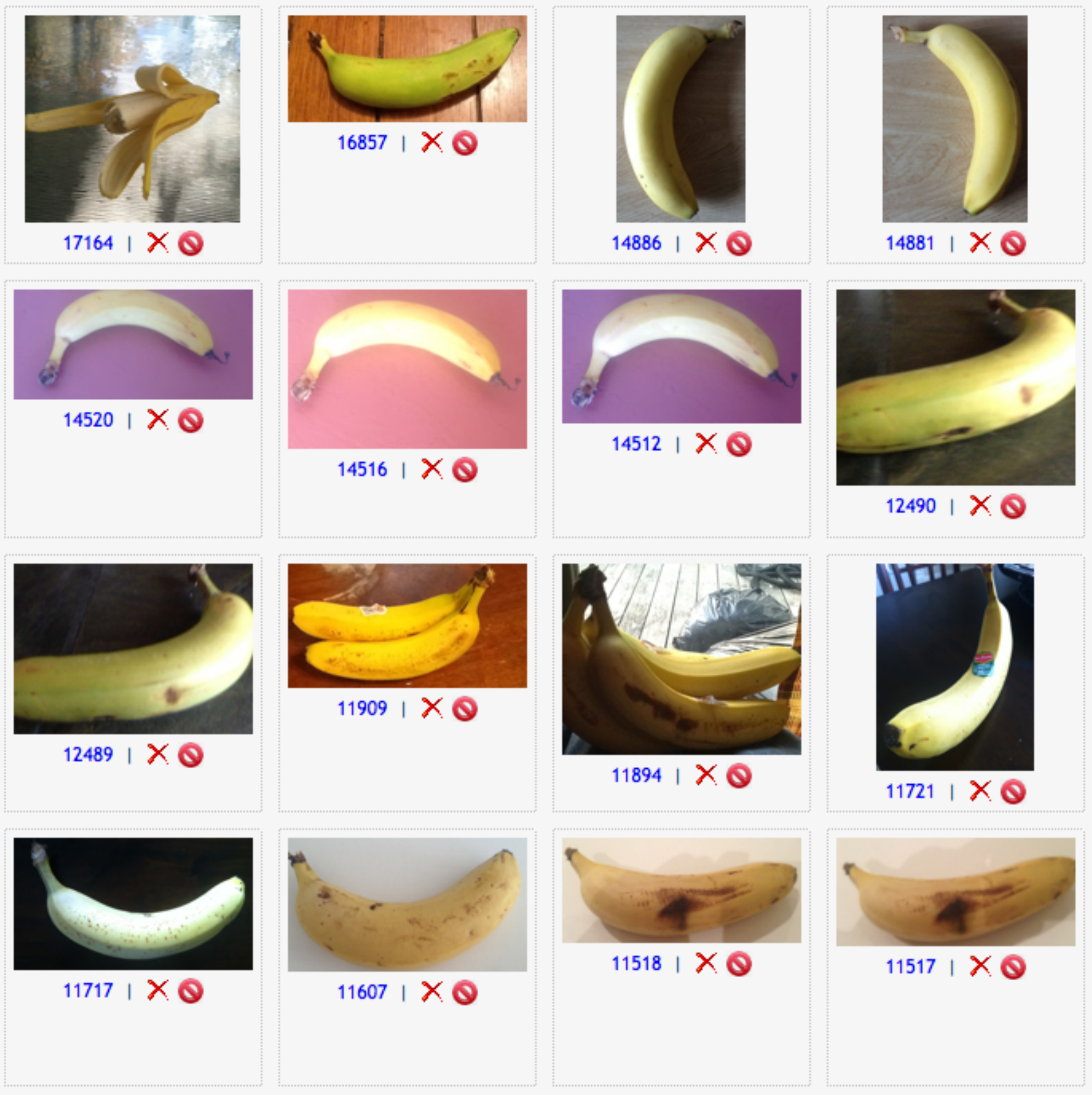} &
    \includegraphics[width=.45\textwidth]{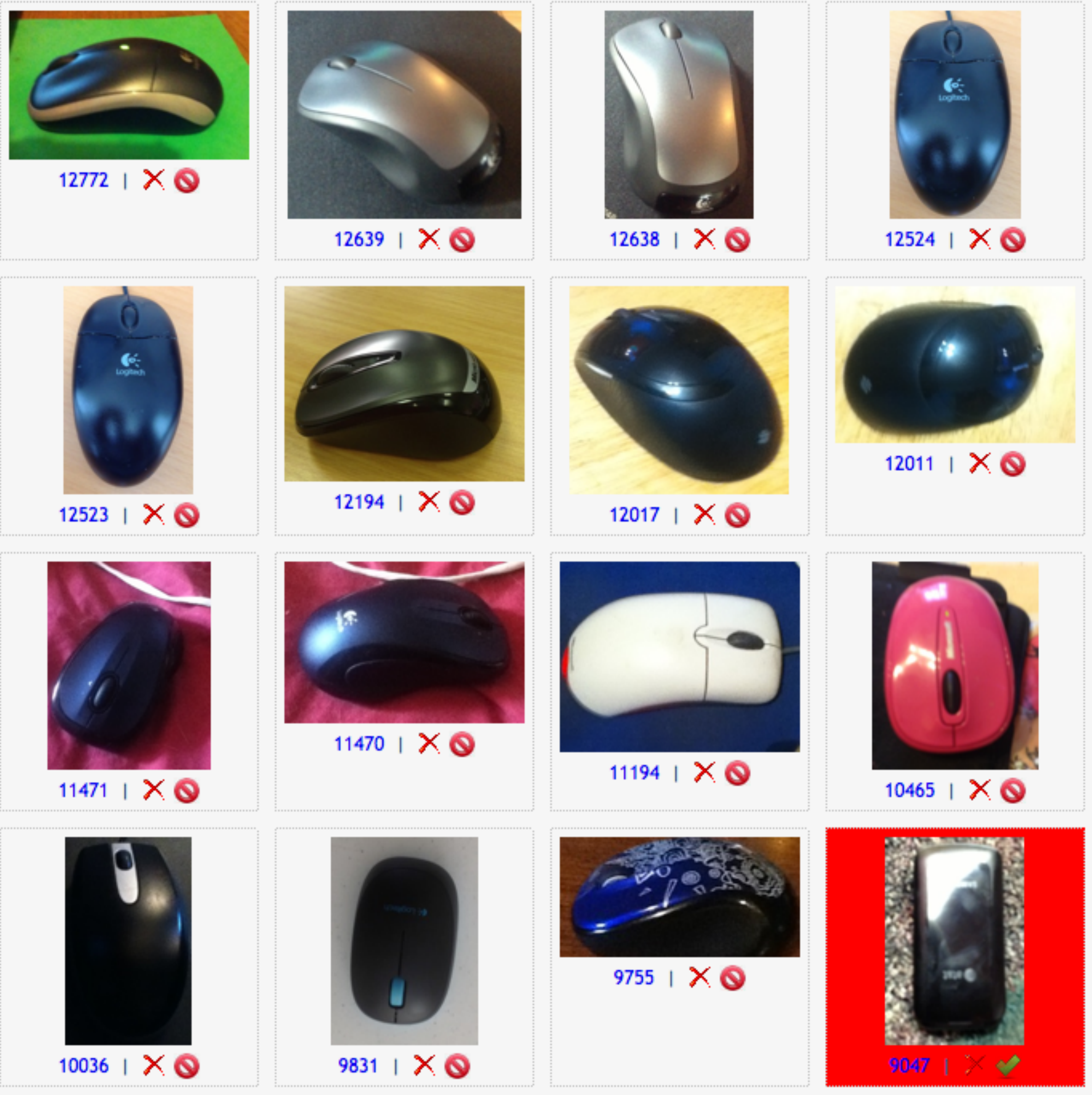} \\
  \end{tabular}
  \caption{\label{FigSamples}Example of images from banana (left) and computer mouse (right) classes. The bottom-right image with a red background is marked as unreliable, as it is a smartphone.}
\end{figure*}

Some examples of the dataset can be seen in Figure \ref{FigSamples}. Images that are out of focus, have a misplaced ROI, or are labeled with a wrong class can be marked as unreliable by the MirBot administrators, who can also delete samples with inappropriate contents.

\begin{figure}
\center
\includegraphics[width=\columnwidth]{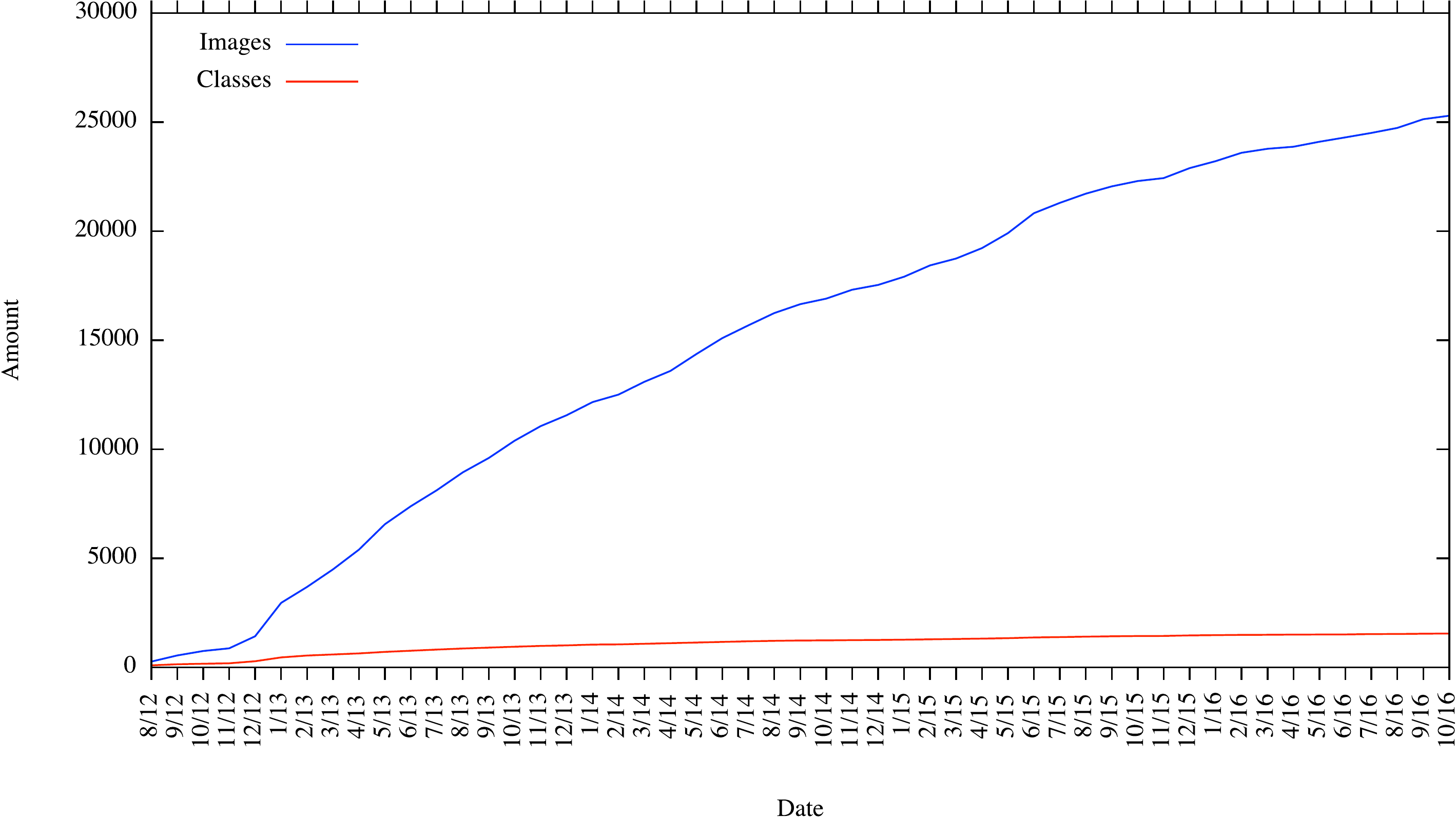}
\caption{\label{Plot1}Evolution  of the number of images and classes over time. }
\end{figure}

As the data is dynamic, the statistics and success rate change over time. The following evaluation results refer to the MirBot dataset on October 23, 2016. On this date, $3,431$ users had added  $25,292$ images distributed in $1,808$ classes. Figure \ref{Plot1} shows the temporal evolution of the data since MirBot has been available.

\begin{figure}[ht]
\center
\includegraphics[width=\columnwidth]{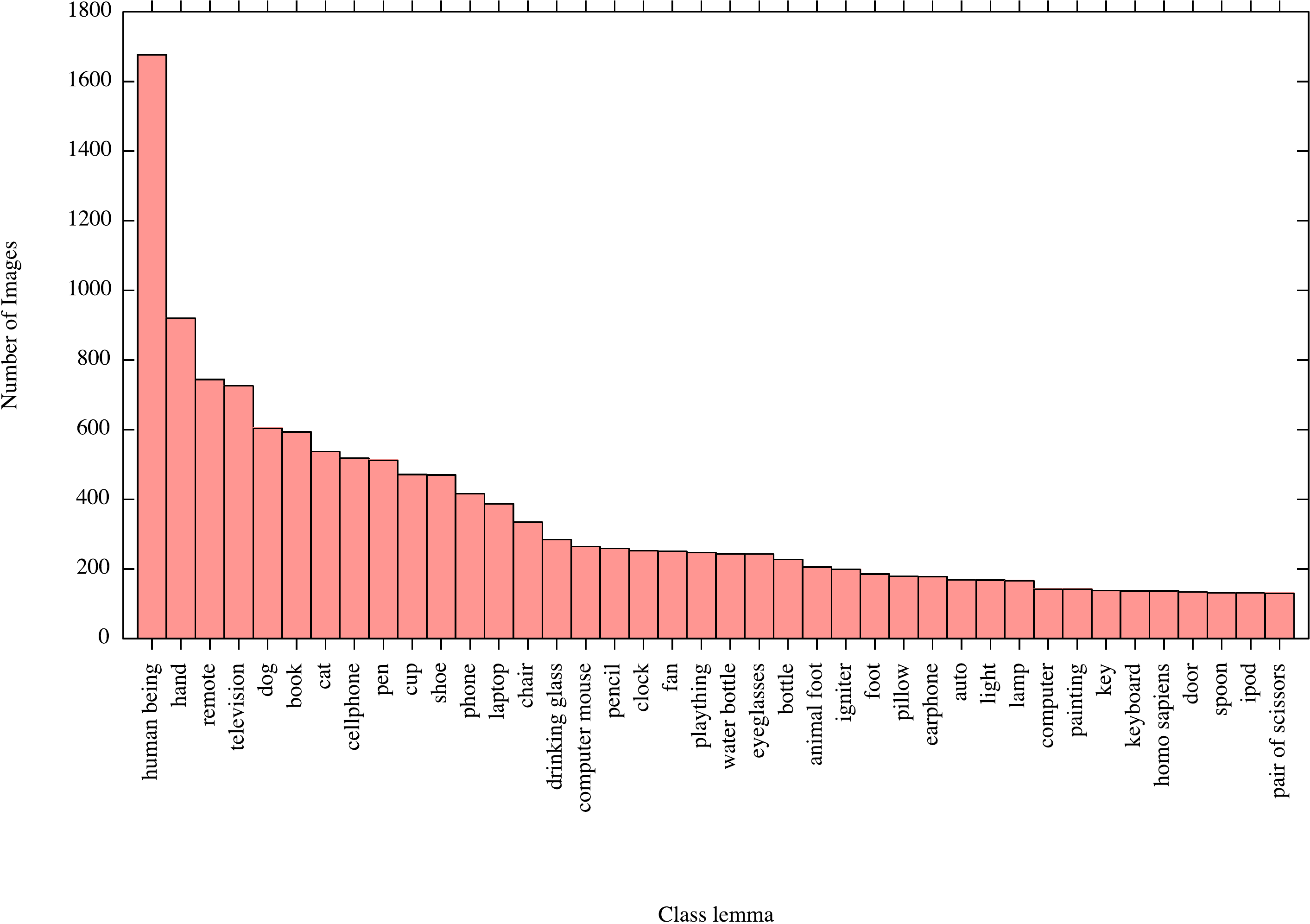}
\caption{\label{Plot2}Number of images for the 40 top classes. }
\end{figure}

Since the dataset is user-driven, some objects appear more frequently than others and the classes are, therefore, unbalanced. Upon observing the WordNet root categories it will be noted that most images are objects ($18,685$), followed by animals ($4,928$), food/drinks ($1,113$), and plants ($546$). Figure \ref{Plot2} shows the top 40 classes by the number of samples. 

The evaluation is performed using a 5-fold cross validation. Only the images belonging to the classes with more than one prototype were used in these experiments ($24,794$ images from $1,180$ classes). 

The accuracy is provided at two levels, top-1 and top-10. In top-1, a true positive is considered when the class of the closest prototype matches the query class. In top-10, a true positive is considered when the query class matches the class of any of its $10$ closest prototypes. Note that this does not mean that $10$ classes are considered, but rather the classes of the $10$ closest prototypes, and if they share the same class only one class is, therefore, taken into account.

\subsection{Results using handcrafted descriptors}

The MirBot base system introduced in \cite{MirBot:System} performed similarity search using only TopSURF descriptors and color histograms. The parameters were experimentally set to $M=64$ (color histogram size), $T=100$ (number of top words in TopSURF), and $D=100,000$ (TopSURF dictionary size). The evaluation results obtained using these values are shown in table \ref{Tab2}.

Some experiments were performed by varying the size of the color histogram $M \in [32,512]$, the number of words $T\in [50,200]$, and the size of the dictionary $D\in [10000, 250000]$. In all these configurations, the success rate varied by a maximum of $0.2$, and these parameters do not, therefore, have a noticeable impact on the accuracy.

In \cite{MirBot:System}, a generic TopSURF dictionary of $100,000$ words was used. An experiment was also performed, which consisted of replacing this dictionary with one trained with the MirBot dataset in order to make the clusters more specific to the target data. The top-1 success rate increased from $14.4$ to $19.6$ when using the trained dictionary, as can be seen in table \ref{Tab2}.

\begin{table}
\centering
\begin{tabular}{l|ll}
\hline
\textbf{Features} 	& \textbf{Top-1} & \textbf{Top-10} \\ \hline
TopSURF generic 	& 14.43 		&35.97 \\
TopSURF trained 	& 19.63 		& 41.10 \\
Color histogram 	& 28.15 		&47.99 \\
Combined (w= 0.1) 	& 36.61 		& 55.95 \\ \hline
VGG16				& 66.64 		&86.56 \\
VGG19				& 67.02 		&86.81 \\
ResNet				& \textbf{71.13} & 88.74 \\
Inception v3		& 65.93 		&85.51 \\
Xception			& 70.52 		& \textbf{88.88} \\
AlexNet				& 48.79 		&76.96 \\
GoogLeNet			& 57.56 		&83.54 \\
Inception21k 		& 61.03 		&86.07 \\
Inception21k-direct & 30.31 		& 75.54 \\
\hline
\end{tabular}
\caption{Top-1 and top-10 average accuracy (in percentages) using a 5-fold cross validation. The handcrafted features evaluated include TopSURF using a generic dictionary with 100,000 words, TopSURF with a trained dictionary of the same size, color YCrCb histograms, and the combined descriptor with color and trained TopSURF with $w=0.1$. Neural codes were extracted from the last hidden layer and normalized using $\ell_2$. Inception21k-direct performs the classification directly using the output layer rather than neural codes.}
\label{Tab2}
\end{table}

Surprisingly, the global color descriptor performed significantly better than the TopSURF local descriptor. The reason for this may be that some pictures from the same object have similar lighting conditions, as users tend to take them consecutively when the system returns a wrong answer. 

We have also evaluated the combination of color and trained TopSURF descriptors (see Eq. \ref{eqWeight}), using the parameter $w$ to balance their contribution. Interestingly, although color outperforms TopSURF, the  combination of both descriptors increases the individual  success rate for some values of $w$. The best results were specifically obtained with $w=0.1$, which was significantly more successful than the color histograms, as can be seen in table \ref{Tab2}.

\subsection{Results using neural codes}

Different parameters can be selected to extract the neural codes from an image. In particular, we have evaluated alternative CNN models, input image scaling techniques, neural activations from different inner layers, and the benefit of normalizing the neural codes using the $\ell_2$ norm before performing the kNN similarity search. 

In our experiments we assessed that normalizing the neural codes using the $\ell_2$ norm consistently improved the accuracy, as can be seen in table \ref{tab:evaluationk}. We also considered different image scaling methods (isotropic or anisotropic rescaling) and alternative input image sizes. As expected, the most adequate input size and rescaling parameters are the same as those originally used to train the networks. The neural codes were extracted from the last hidden layer, but we also carried out experiments using different layers. The last hidden layer obtained the best results in all cases and was, therefore, selected for the following experiments. 

Neural codes from different topologies (detailed in Section \ref{sec:ncpm}) were also evaluated, and clearly outperformed the handcrafted features, as shown in table \ref{Tab2}. MirBot initially used the combined Top-Surf and color descriptor. However, GoogLeNet replaced the handcrafted descriptors in June 2015, and from October 2016 to the present the features have been extracted with Inception21k, although given the evaluation results we plan to replace them with ResNet soon.

Table \ref{Tab2} shows the 5-fold cross validation average accuracy using the weights of the models trained with the ImageNet 1,000 classes subset (ILSVRC12), with the exception of Inception21k, which was trained with the full ImageNet dataset.

As the evaluation results show, the pretrained models obtain high success rates, especially when taking into account that the target classes are different from those used for training. The best accuracies are obtained with ResNet and Xception, which are those that yielded the highest accuracies in the ILSVRC contest.

Experiments were also performed by varying the value of $k$. Table \ref{tab:evaluationk} shows the results obtained with and without the $\ell_2$ normalization of the neural codes. As can be seen, the highest accuracies are obtained with $k=1$ and normalized codes. This low value of $k$ could be owing to the highly unbalanced dataset, as some classes contain very few samples. 

Besides transfer learning, we have also used the SoftMax output layer of the Inception21k network (namely Inception21k-direct), as this CNN was originally trained using the full ImageNet dataset and it therefore contains most of the classes that are present in MirBot. In these experiments we considered only the activations of the output neurons belonging to the MirBot classes that have more than one prototype. In this case, only $20,400$ images were used for evaluation, as $213$ of the MirBot classes were not present in the $21,841$ ImageNet classes. These results cannot be directly compared to the transfer learning approaches because the dataset is different, but we can obtain an approximate success rate. 

The top-1 accuracy ($30.31$) in Inception21k-direct is much lower than that from the transfer learning representations, whereas the top-10 ($75.54$) is also lower but not to such a great extent.  This discrepancy can be explained by the fact that some images may fit in different classes, signifying that these classes share a very similar content. For example, an image of a person may be classified either as a human-being, male, adult, or even homo-sapiens-sapiens. 

\begin{table*}
\centering
\begin{footnotesize}
\begin{tabular}{llcccccccc}
\hline
&& \multicolumn{2}{c}{\textbf{$k$=1}} & \multicolumn{1}{l}{} & \multicolumn{2}{c}{\textbf{$k$=5}} & \multicolumn{1}{l}{} & \multicolumn{2}{c}{\textbf{$k$=10}} \\ \cline{3-4} \cline{6-7} \cline{9-10} 
\textbf{Network} && -         & $\ell_2$           && -         & $\ell_2$           && -          & $\ell_2$           \\ \cline{1-1} \cline{3-4} \cline{6-7} \cline{9-10} 
\textbf{VGG16}                                                    && 63.00     & 66.64              && 57.08     & 60.94              && 55.04      & 58.92              \\ 
\textbf{VGG19}                                                    && 63.56     & 67.02              && 57.53     & 61.07              && 54.93      & 59.12              \\ 
\textbf{ResNet}                                                    && \textbf{69.77}     & \textbf{71.13}     && \textbf{62.70}     & 64.73              && 60.54      & 62.29              \\ 
\textbf{Inception v3}                                              && 67.12     & 65.93              && 61.39     & 59.91              && 58.86      & 57.86              \\ 
\textbf{Xception}                                                  && 68.68     & 70.52              && 62.54     & \textbf{65.15}     && \textbf{60.88}      & \textbf{63.33}     \\ 
\textbf{AlexNet}                                                   && 46.31     & 48.79              && 45.99     & 48.45              && 46.29      & 48.53              \\ 
\textbf{GoogLeNet}                                                 && 56.87     & 57.56              && 54.94     & 56.48              && 54.20      & 56.16              \\ 
\textbf{Inception21k}                                              && 60.02     & 61.03              && 58.98     & 60.33              && 58.37      & 59.92              \\ \hline
\end{tabular}
\end{footnotesize}
\caption{Top-1 accuracy using 5-fold cross-validation with and without $\ell_2$ for different values of $k$.}
\label{tab:evaluationk}
\end{table*}

\subsection{Results with fine tuning and full training}

Besides using the ImageNet pretrained weights, we have also trained the best 5 CNN models (according to the previous results) with the MirBot dataset. This training was performed using a maximum of 500 epochs, a mini-batch size of 16 samples, and an early stopping of 10 epochs. In the fine tuning experiments, the original fully-connected (FC) layers were replaced with three new FC layers in which the last one had the same number of neurons as the amount of MirBot classes (1,256) that contained more than one sample. We have evaluated three learning approaches:

\begin{itemize}
\item Full training: The model is trained from scratch. 
\item Fine tuning last layers: Fine-tuning using the weights of convolutional layers from ImageNet and replacing the last fully-connected layers with 3 fully-connected layers whose weights must be learned. 
\item Fine tuning from the middle: Fine-tuning starting from the middle layer of the model whose weights were originally trained using ImageNet. The use of this scheme makes it possible to adjust more network weights for the new classes than in the previous approach.
\end{itemize}

The evaluation results obtained using the SoftMax activations of the trained models can be seen in Table \ref{tab:dnn_train}. Full-training accuracy is very low because the number of samples in the MirBot dataset is not sufficient to train models that contain a large number of parameters and because the classes are highly unbalanced. The best results are obtained by fine-tuning only the last layers, although the accuracy is still below that obtained using the pretrained features and kNN.

In order to assess the accuracy in the same scenario employed with the pretrained models, experiments were also performed by replacing the SoftMax layer with a kNN after full training or fine tuning. This was done by extracting neural codes from the last hidden layer, normalizing them using $\ell_2$, and then returning the class of the closest prototype ($k=1$) in the training set. The results are shown in Table \ref{tab:knn_from_dnn_train}. After testing different values, the size of the neural codes was set to 1,256, with the exception of the first column in this table (without training), in which they have the size of the last hidden layer from the original model. 

It will be noted that, when fine-tuning the pretrained ILSVRC models starting from the middle layers and then use the normalized neural codes from the last hidden layer in the prediction stage, the accuracy outperforms the original pretrained models. Moreover, in all cases the accuracy obtained using the SoftMax output is clearly below that attained when using neural codes with kNN in the prediction stage. 

\begin{table}
\renewcommand{\arraystretch}{1}
\centering
\begin{footnotesize}
\begin{tabular}{lccc}
\hline
\textbf{Network}      & \textbf{\begin{tabular}[c]{@{}c@{}}full \\ training\end{tabular}} & \textbf{\begin{tabular}[c]{@{}c@{}}f. tune from \\ the middle\end{tabular}} & \textbf{\begin{tabular}[c]{@{}c@{}}f. tune \\ last layers\end{tabular}} \\ \hline
\textbf{VGG16}        & 22.28                & 32.32                                                                     & 56.91                                                                 \\ 
\textbf{VGG19}        & 21.15                & 30.24                                                                     & 56.83                                                                 \\ 
\textbf{ResNet}       & 19.91                & 60.16                                                                     & \textbf{61.54}                                                        \\ 
\textbf{Inception v3} & 12.36                & 43.61                                                                     & 49.70                                                                 \\ 
\textbf{Xception}     & 10.82                & 39.53                                                                     & 12.32                                                                 \\ \hline
\end{tabular}
\end{footnotesize}
\caption{Top-1 accuracy using the SoftMax output (without kNN) for the different models and training strategies evaluated.}
\label{tab:dnn_train}
\end{table}

\begin{table*}
\setlength{\tabcolsep}{2.5pt}
\renewcommand{\arraystretch}{1}
\centering
\begin{footnotesize}
\begin{tabular}{llccccccccccc}
\hline
&& \multicolumn{2}{c}{\textbf{\begin{tabular}[c]{@{}c@{}}Without \\ training\end{tabular}}} & \textbf{} & \multicolumn{2}{c}{\textbf{Full training}} & \textbf{} & \multicolumn{2}{c}{\textbf{\begin{tabular}[c]{@{}c@{}}F. tune from \\ the middle\end{tabular}}} & \textbf{} & \multicolumn{2}{c}{\textbf{\begin{tabular}[c]{@{}c@{}}F. tune \\ last layers\end{tabular}}} \\ \cline{3-4} \cline{6-7} \cline{9-10} \cline{12-13} 
\textbf{Network}      && \textbf{Top-1}                              & \textbf{Top-10}                            & \textbf{} & \textbf{Top-1}       & \textbf{Top-10}     & \textbf{} & \textbf{Top-1}                                 & \textbf{Top-10}                                & \textbf{} & \textbf{Top-1}                               & \textbf{Top-10}                              \\ \cline{1-1} \cline{3-4} \cline{6-7} \cline{9-10} \cline{12-13} 
\textbf{VGG16}        && 66.64                                     & 86.56                                    && \textbf{38.99}     & \textbf{63.80}    && 47.60                                        & 69.60                                        && 61.51                                      & 73.14                                      \\ 
\textbf{VGG19}        && 67.02                                     & 86.81                                    && 34.32              & 59.40             && 43.84                                        & 67.19                                        && \textbf{75.86}                             & 89.51                                      \\ 
\textbf{ResNet}       && \textbf{71.13}                            & 88.74                                    && 37.81              & 61.44             && \textbf{77.70}                               & \textbf{90.68}                               && 73.51                                      & \textbf{89.70}                             \\ 
\textbf{Inception v3} && 65.93                                     & 85.51                                    && 22.64              & 46.76             && 60.94                                        & 78.88                                        && 67.35                                      & 83.02                                      \\ 
\textbf{Xception}     && 70.52                                     & \textbf{88.88}                           && 25.31              & 45.11             && 58.83                                        & 78.14                                        && 65.65                                      & 81.26                                      \\ \hline
\textbf{Average}      && 68.25                                     & 87.30                                    && 31.81              & 55.30             && 57.78                                        & 76.90                                        && 68.78                                      & 83.33                                        \\ \hline
\end{tabular}
\end{footnotesize}
\caption{Top-1 and top-10 results with the fully-trained or fine-tuned models using $\ell_2$ normalized vector codes extracted from the last hidden layer. 5-fold cross-validation experiments were performed in the same way as in Table \ref{Tab2}.}
\label{tab:knn_from_dnn_train}
\end{table*}

\subsection{Accuracy over time}

\begin{figure}
\center
\includegraphics[width=\columnwidth]{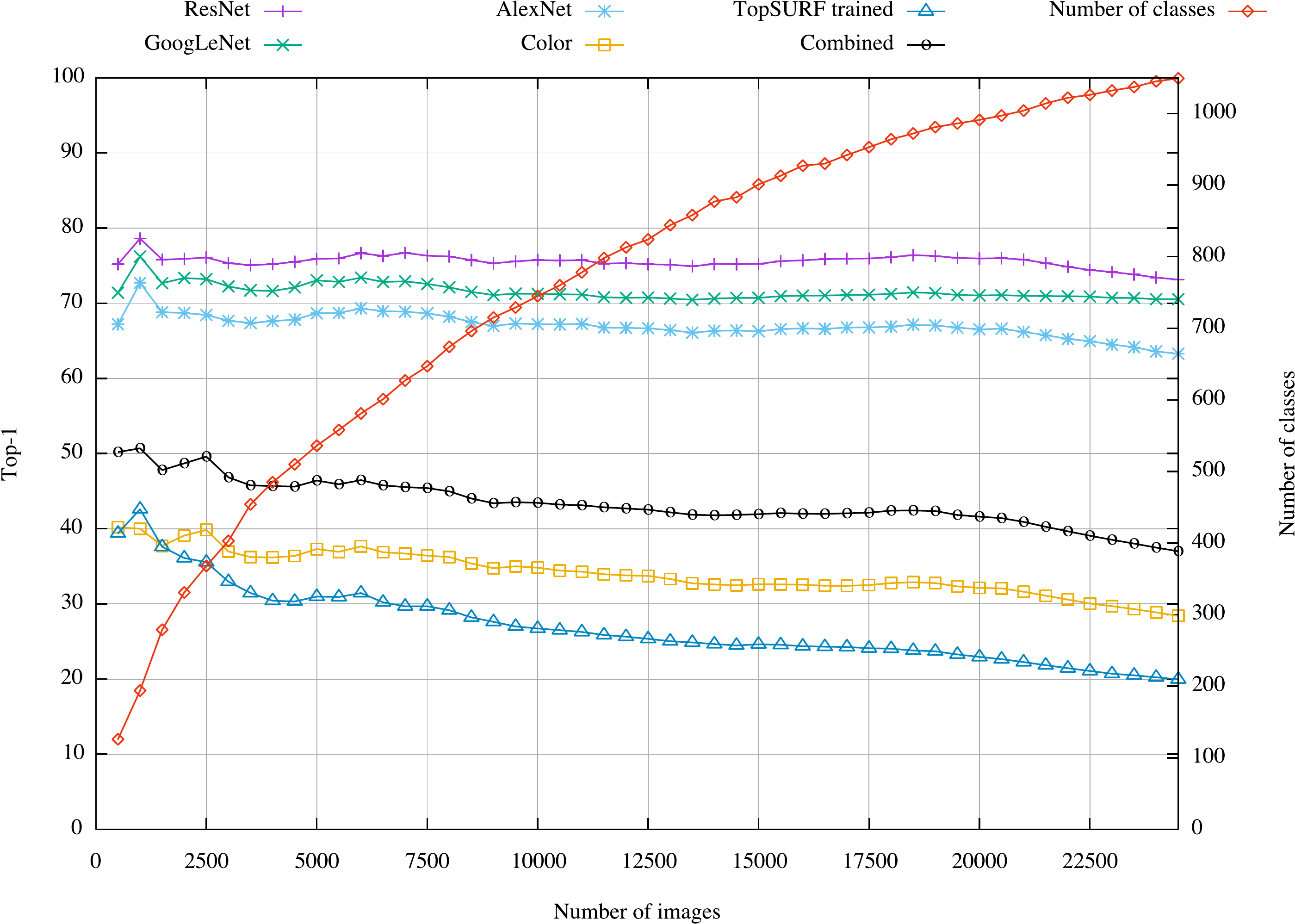}
\caption{\label{Plot4}Number of classes and temporal evolution of the top-1 success rate for the evaluated descriptors when increasing the number of images. A leaving-one-out experiment was subsequently performed for the first $500$ images of the database, then for the $1,000$ first images, and so on, until the full database had been evaluated. The combined descriptor using TopSURF and color histogram was evaluated with $w=0.1$. Although the number of classes increases with the number of images, the success rate remained almost constant with the CNN representations unlike it happened using handcrafted features (ResNet accuracy oscillated between $76.3$ and $73.2$, whereas the combined descriptor accuracy decreased from $50.2$ to $37$).}
\end{figure}

We have also evaluated the handcrafted descriptors and the neural codes with regard to the number of images (see Figure \ref{Plot4}). In this case, we performed top-1 leaving-one-out experiments with the first $500$ images that were added to the dataset, and then with the first $1,000$ images, until all the images had been considered. The goal of this experiment was to check the evolution of the MirBot success rate over time. For ease of viewing, only three representative CNN models pretrained with ILSVRC12 data are shown in Fig. \ref{Plot4}. 

The results of this experiment are very interesting. After the initial spikes owing to the small number of samples, the neural codes maintain an almost constant success rate despite the increasing number of classes. The main reason for this is that there are more classes over time but also more images in the database with which to compare them, thus signifying that the success rate over time remains constant with these descriptors. The accuracy for handcrafted descriptors, however, significantly decreased over time, showing that they are less robust.

\subsection{Accuracy attained when increasing the minimum number of samples per class.}

\begin{figure}
\center
\includegraphics[width=\columnwidth]{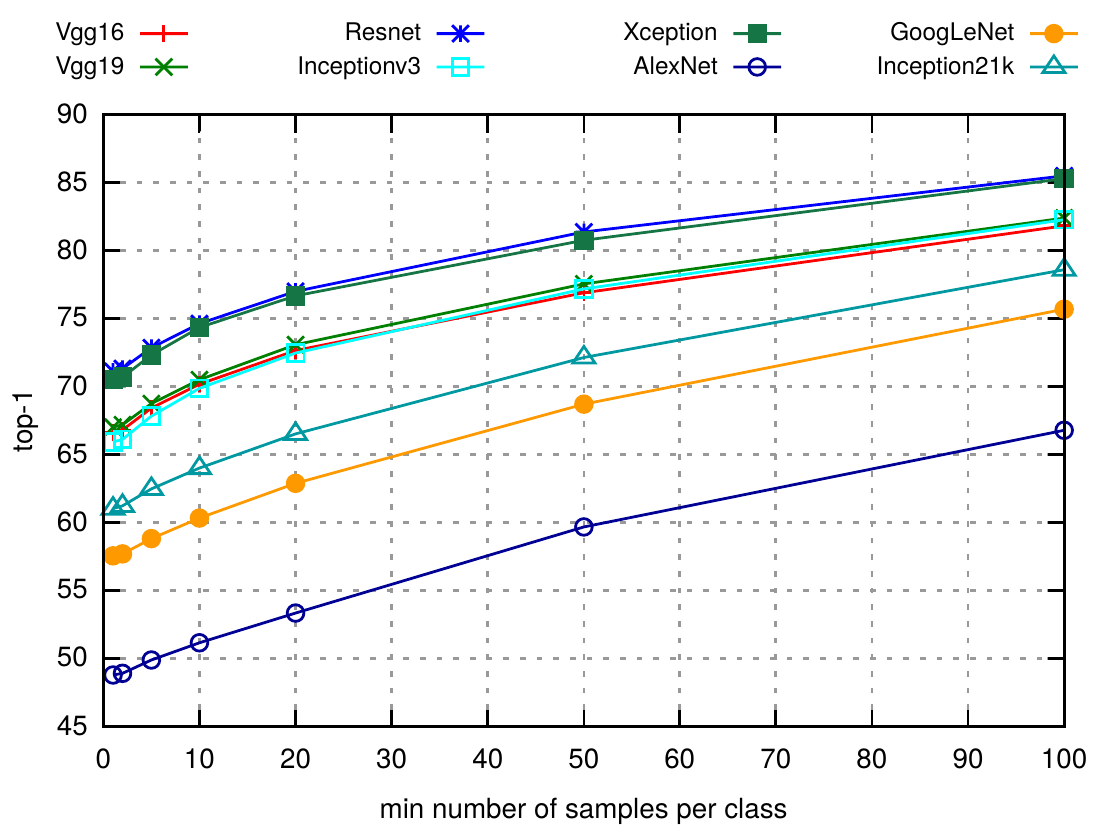}
\caption{\label{fig:plot_min_samples}Top-1 accuracy attained from 5-fold cross validation using only those classes with a minimum number of samples.}
\end{figure}

A similar experiment was performed with regard to the number of samples per class, during which the accuracy was evaluated using those classes which contain a minimum number of samples (from 1 to 100). The first column in figure \ref{fig:plot_min_samples} shows the top-1 accuracy using all classes, while the last one shows the accuracy attained when using only those classes with a minimum number of 100 samples. The experiments were performed with a 5-fold cross validation using the best parameters found in the previous evaluation (neural codes from the last hidden layer, $\ell_2$ norm and $k=1$). 

It will be noted that the top-1 accuracy increases when more samples per class are available. As the number of new classes converges (as shown in Figure \ref{Plot4}) and new images are continuously added, the accuracy is expected to increase over time as the MirBot dataset grows. 

\subsection{Evaluation using PCA compression}
\label{sec:pca}

The neural codes can be compressed using Principal Component Analysis (PCA). In \cite{BabenkoSCL14} it is observed that their dimensionality can be reduced very substantially, e.g. to 128 dimensions with virtually no accuracy loss. We have also evaluated the performance of PCA to reduce the dimensionality of the neural codes in the MirBot dataset with the different network topologies.


PCA performs a linear dimensionality reduction using Singular Value Decomposition (SVD) of the data to project it to a lower dimensional space. For the following experiments we used the LAPACK implementation of the full SVD to automatically select the number of components such that the amount of variance that needs to be explained is greater than 0.95.


 
As previously mentioned, the dimensionality of the neural codes in our trained networks is 1256, whereas the pretrained networks have a size of 4096 for VGGs and AlexNet, and 2048 for ResNet, Inception v3 and Xception. 

Table \ref{tab:pca-components} shows the new dimensionality obtained for each type of network and training strategy after applying PCA compression with a threshold of 0.95. 
Overall, the dimensionality is reduced from about  50\% without training to 90\% when fine-tuning from the middle.
%

\begin{table*}
\centering
\footnotesize

\begin{tabular}{lcccc}
\hline
\textbf{Network}      & \textbf{\begin{tabular}[c]{@{}c@{}}Without\\training\end{tabular}} & \textbf{\begin{tabular}[c]{@{}c@{}}Full\\training\end{tabular}} & \textbf{\begin{tabular}[c]{@{}c@{}}F. tune from\\the middle\end{tabular}} & \textbf{\begin{tabular}[c]{@{}c@{}}F. tune\\last layers\end{tabular}} \\ \hline
\textbf{AlexNet}	  & 2175		 & 527					  & 531			 & 644 \\
\textbf{VGG16}        & 1980         & 686                    & 116          & 655 \\
\textbf{VGG19}        & 1939         & 332                    & 22           & 728 \\
\textbf{ResNet}       & 1036         & 174                    & 307          & 601 \\
\textbf{Inception v3} & 776          & 514                    & 41           & 88  \\
\textbf{Xception}     & 1258         & 195                    & 124          & 71  \\ 
\hline
\textbf{Average}      & 1527         & 405                    & 190          & 464 \\ 
\hline
\end{tabular}

\caption{Number of PCA components using a variance threshold of 0.95.}
\label{tab:pca-components}
\end{table*}

Table \ref{tab:pca_results} shows the resulting performance using PCA compression with the new dimensionality in each case. It can be seen that PCA introduces an accuracy loss. The best score with ResNet decreases from 77.70 to 71.02, whereas other models such as VGG19 decrease less (from 75.86 to 72.82).  

\begin{table*}
\centering
\footnotesize
\setlength{\tabcolsep}{2.5pt}
\renewcommand{\arraystretch}{1}

\begin{tabular}{llccccccccccc}
\hline
             &  & \multicolumn{2}{c}{\textbf{\begin{tabular}[c]{@{}c@{}}Without\\ training\end{tabular}}} &  & \multicolumn{2}{c}{\textbf{Full training}} &  & \multicolumn{2}{c}{\textbf{\begin{tabular}[c]{@{}c@{}}F. tune from \\ the middle\end{tabular}}} &  & \multicolumn{2}{c}{\textbf{\begin{tabular}[c]{@{}c@{}}F. tune \\ last layers\end{tabular}}} \\ 
             \cline{1-1} \cline{3-4} \cline{6-7} \cline{9-10} \cline{12-13} 
\textbf{Network}      &  & \textbf{NC}    & \textbf{PCA}     &  & \textbf{NC}    & \textbf{PCA}     &  & \textbf{NC}    & \textbf{PCA}    &  & \textbf{NC}   & \textbf{PCA}     \\ 
	\cline{1-1} \cline{3-4} \cline{6-7} \cline{9-10} \cline{12-13} 
\textbf{AlexNet}      &  & 48.79	       & 	 46.65         &  & 	36.74          & 	32.83        &  & 32.85 & 29.14	      &  & 53.53 & 50.23           \\
\textbf{VGG16}        &  & 66.64 & 63.78         &  & \textbf{38.99} & 35.81   &  & 47.60 & 44.38       &  & 61.51 & 45.74            \\
\textbf{VGG19}        &  & 67.02 & 64.27          &  & 34.32 & 32.55           &  & 43.84 & 32.99       &  & \textbf{75.86} & \textbf{72.82}  \\
\textbf{ResNet}       &  &\textbf{71.13} &  67.75           &  & 37.81 & \textbf{38.68}            &  & \textbf{77.70} & \textbf{71.02}  &  & 73.51 & 68.44                   \\
\textbf{Inception v3} &  & 65.93 & 62.72          &  & 22.64 & 18.33           &  & 60.94 & 47.37         &  & 67.35 & 56.97                    \\
\textbf{Xception}     &  & 70.52 & \textbf{67.77}      &  & 25.31 &  28.35                  &  & 58.83 & 59.77            &  & 65.65 &  55.48             \\ 
\hline
\textbf{Average}      &  & 65.01 & 62.16	     &  & 32.64 & 31.09	        	        &  & 53.63 & 47.45	           &  & 66.23 & 58.28	            \\
\hline
\end{tabular}

\caption{Top-1 results for uncompressed neural codes and PCA using all the networks and training schemes evaluated.}
\label{tab:pca_results}
\end{table*}

These performance losses are a bit higher than those obtained in \cite{BabenkoSCL14}, where virtually the same results were reported with PCA. One explanation for this is that only one network -- AlexNet \cite{Krizhevsky2012ImageNetNetworks} -- was evaluated in \cite{BabenkoSCL14}. The weights of this network can be easily compressed as they are very sparse \cite{Sparse}, therefore  the accuracy loss is smaller when using PCA on these NC. However, recent networks such as ResNet are much less sparse, therefore their PCA loss is larger as it can be seen in Table \ref{tab:pca_results}. 

Although accuracy decreases with PCA, it must be considered these results are obtained using vectors with a fraction of their original size. For example, the ResNet neural codes with fine tuning from the middle are compressed to 307 dimensions with a loss of only 6 points.

The computational cost using PCA and Annoy is shown in Table \ref{tab:pca_times}. PCA is significantly faster (12.3 times) than the original NC. However, it is slower than the approximate nearest neighbors method (Annoy) evaluated, which has a negligible accuracy loss but at a larger memory footprint. Obviously PCA compressed codes could also be searched with Annoy, and in this case the computational cost is marginal. In MirBot we are currently using the original neural codes with Annoy, but PCA with Annoy will be considered if the number of images increases significantly. 

\begin{table*}
\centering
\footnotesize

\begin{tabular}{lcccc}
\hline
\textbf{Network}      & \textbf{NC} & \textbf{PCA} & \textbf{NC+Annoy} & \textbf{PCA+Annoy} \\ \hline
\textbf{AlexNet}      & 180.3  & 25.2  & 7.8 & 3.37  \\ 
\textbf{VGG16}        & 122.4  & 10.5   & 4.6 & 0.95          \\
\textbf{VGG19}        & 121.3  & 9.0    & 4.7 & 0.54          \\
\textbf{ResNet}       & 60.0  & 6.8     & 2.4 & 0.48            \\
\textbf{Inception v3} & 67.6   & 4.1    & 2.4 & 0.78         \\
\textbf{Xception}     & 60.5    & 4.5 &  2.2  & 0.45         \\ \hline
\textbf{Average}      & 86.4   & 7.0   &  3.26  & 0.64    \\ \hline
\end{tabular}

\caption{Comparison of the average time in milliseconds to search a query with NC, compressed PCA (brute-force, 4 threads), Annoy (single thread) with the original NC, and Annoy (single thread) with the PCA compressed codes.}
\label{tab:pca_times}
\end{table*}

\subsection{Metadata evaluation}
\label{sec:metadata}

As previously mentioned, the MirBot dataset also includes a series of metadata, described in \cite{MirBot:System}. They include information from the smartphone sensors, reverse geocoding and EXIF camera data.
The metadata from the sensors correspond to the device information, geolocation data, and the accelerometer, gyroscope, and network. In addition, given a latitude and a longitude, reverse geocoding is performed in the server with Gisgraphy2, which uses the Geo-Names geographical database to obtain relevant data such as the feature class and code that provide information about the kind of place. The parameters of the photographs are also stored using the  exchangeable image file format (EXIF), which includes 23 features such as the aperture value, brightness, ISO speed, white balance, etc. The complete list of the 79 metadata stored in MirBot can be seen in \cite{MirBot:System}.

In order to evaluate the classification performance using only these metadata, the features \textit{osversion} and \textit{model} have been removed, along with all the information related to an specific user such as its identifier. 

Those metadata containing real or integer values (such as pitch, sharpness, focal length, etc.) were normalized in the interval $[0,1]$. However, categorical metadata (such as country, gis feature code, gis feature class, etc.) were codified in a one-hot manner as they have not any specific ordering. This way, the distance between two categorical features can only be 1 if they are different or 0 if they match.

First, attribute selection have been performed in order to rank the metadata and select the best subset. For this we have applied several selection methods~\cite{weka}: 
Best first, Genetic search, Greedy Stepwise, Linear Forward Selection, Random Search, Scatter Search V1, Subset Size Forward Selection, and InfoGain. 
After testing all these methods we applied a voting scheme in order to select the best attributes. The most voted attributes by the selection methods are shown in Table \ref{tab:metadata-selected}, and they were chosen for the next evaluation stage, ignoring the others. 


\begin{table}
\centering
\footnotesize
\begin{tabular}{lll}
\hline
\textbf{Sensors} & \textbf{Location}  & \textbf{EXIF}    \\ \hline
pitch           & reliable location  & sharpness        \\
selected area   & country            & focal length     \\
wifi            & ocean              & brightness value \\
flash           & gis feature code   & color space      \\
                & gis feature class  & subject area     \\
\hline
\end{tabular}
\caption{Most representative metadata using different attribute selection methods with a voting scheme.}
\label{tab:metadata-selected}
\end{table}

One of the most representative metadata is the feature code \cite{feat_codes}, which stores the kind of place: Zoo, Mall, University, Beach, etc. On the other hand, others such as the angle, which we expected to perform well in the original MirBot version \cite{MirBot:System}, were not selected.

We performed an initial evaluation using only this metadata subset without any image features. The classification was done at three levels: Root level (with the 5 main categories: animals, food and drink, man-made objects, natural objects, and plants), the second level of the WordNet hierarchy (with 92 classes), and the leaf level (with the 1119 classes). 

Three classifiers were evaluated for this task: kNN with $k \in [1,100]$; Support Vector Machines (SVM) with $C \in [1,1000]$; and Random Forests (RF), with the number of trees within the range $[5,1000]$.  
 
Table \ref{tab:metadata-results} shows the best results for each classifier using 5-fold cross-validation. The best results with kNN were obtained with a very high value ($k=80$), the best results using RF were with 150-300 trees, and with SVM the accuracy did not improve with values of $C$ larger than $10$.

The results obtained for the first levels of the hierarchy are surprisingly high considering that the classification is performed without any visual information. An explanation for this is that the dataset is unbalanced, therefore the baseline is also high, although it is lower than the reported results. We have checked the confusion matrices in order to assess that the yielded classes are varied and there is no overfitting.

\begin{table}
\centering
\footnotesize
\begin{tabular}{lccc}
\hline
                & \multicolumn{3}{c}{\textbf{Classification type}}    \\ \hline
\textbf{Method} & \textbf{Root} & \textbf{2nd level} & \textbf{Class} \\ \hline
\textbf{kNN}    & 73.67         & 51.73              & 7.31           \\
\textbf{RF}     & 67.8          & 35.31              & 9.94           \\
\textbf{SVM}    & 73.7          & 52.01              & 6.29           \\ \hline
\end{tabular}
\caption{Best results for each classifier using only the metadata.}
\label{tab:metadata-results}
\end{table}

Additional experiments were made by combining the  metadata with the neural codes from the images. For this, both early and late fusion were performed. In the case of late-fusion, the NC vector was concatenated to the metadata vector. Then, kNN, SVM and RF were used on these data. For early-fusion, an initial classification was performed using the metadata at the root level. Then only the NC codes from that category were considered to be classified  with kNN, SVM and RF. As the accuracy with metadata is much lower than the one obtained with neural codes, the overall accuracy decreased in both early and late fusion around a 6\%.

A third approach was tested: given that the neural codes classifier is much more reliable, we only use metadata when its confidence is low, that is when the difference of the distances between the first and second class returned is small. For the following experiment we have used only one metadata, the feature code, which experimentally was the most significant. 

The algorithm is as follows: First we build a normalized histogram of feature codes for each class in the training set. Then, we classify a query using ResNet. If the difference between the distances of the most likely class and the second class returned is smaller than a threshold $\rho$, we assume a low image classifier confidence. In that case, the feature code of the query is checked in the histograms of both classes. If that feature code has a larger value in the histogram of the second class than in the first one, then we return it as the most likely class, changing this way the class order returned by the image classifier.

We evaluated this setup using a threshold $\rho = 0.02$, which is the average distance between the first and second classes when a wrong result is returned by the NC  classifier. Using this threshold value, a 9.66\% of the samples were reranked with metadata. The 70.1\% of these changes did not have effect as the correct class was neither the first or the second one returned. In a 21.82 \% of these changes the wrong class was properly fixed, and only in a 4.29 \% a correct class switched by a wrong one. Overall, the Top-1 accuracy using pretrained ResNet improved from 71.13 to 74.16. 

As can be seen, a simple methodology using just one metadata (the feature code) has  improved the accuracy in a two-stage setup. Although these results are promising, evaluation should be thoroughly explored with additional metadata.

\section{Conclusions and future work}
\label{conclusions}

This paper presents a description and evaluation of an interactive image recognition system for smartphones. The proposed method, which is freely available~\footnote{\url{http://www.mirbot.com}} for iOS and Android devices, allows a user to train an object recognition system using the classes from a semantic dictionary. 

One of the main research results of this work is a multimodal dataset of images taken with smartphones and organized according to the WordNet ontology. This dataset is continuously growing as it is user-driven. Its images are labeled and stored with a series of metadata, within a ROI, with minimum occlusions, and usually with plain backgrounds, as they are specifically taken for this task and are not downloaded from the Internet. A web interface allows researchers to explore, review and download images, metadata and descriptors. 

An interaction methodology is also presented, which allows users to easily set the image class in a few steps, adding dynamic messages in order to make the labeling process enjoyable. Besides setting the class of the images, users can also train the system to identify objects in their collections (such as stamps, butterflies, fossils, etc.) using the optional labels. However, most users treat the app as a casual game rather than a tool. Interestingly, our most frequent users are children, perhaps because they like the idea of teaching a robot.

Evaluation has been performed using top-1 and top-10 accuracy with different visual descriptors in order to obtain a baseline for future studies. The performance of color histograms, local features and different convolutional neural network representations is assessed. 

The results obtained using handcrafted descriptors show that the combination of color and local descriptors may increase the success rate  when compared to using them independently. However, CNN features outperform these descriptors by far, and their accuracy does not decrease with regard to the number of images when the database grows, unlike that which occurs with the handcrafted features.  The usage of neural codes and kNN rather than the SoftMax output layer allows the system to be incremental when new classes are added. 

As can be seen in the evaluation section, using ImageNet pretrained models to extract the neural codes yields a high accuracy even when the MirBot classes are different from those used to train the model. We also evaluated different fine-tuning strategies in order to adapt the weights to the data collected in MirBot. The best results were obtained using fine-tuning from the middle layers of a ResNet network, which was more accurate than the pretrained models. Moreover, we can conclude that the neural codes and kNN clearly provide more accuracy than the common prediction strategy (using the SoftMax output), as is shown in tables \ref{tab:dnn_train} and \ref{tab:knn_from_dnn_train}. We also verified that adding the $\ell_2$ norm improves the accuracy in all the scenarios tested. The efficiency and performance of PCA compression is also evaluated with the NC of the different networks. 


Metadata can complement the visual descriptors using different multimodal techniques, such as early fusion, late fusion, or with a joint approach (using multimodal neural networks in a manner similar to that of \cite{Mao2014ExplainNetworks}). Initial early and late fusion experiments were performed, showing an accuracy decrease of the joint method due to the low accuracy of the metadata classifier. In order to avoid this, similarity metric learning techniques \citep{Bellet2012SimilarityClassification} could be used in a future work to give less importance to metadata. However, a two-step methodology has shown to be adequate; when the image classifier confidence is small, then metadata is used to refine the results. A preliminary evaluation using only one metadata (the type of location) shows a significant accuracy improvement. 

As the dataset is incremental, the classes are unbalanced, which may be an issue for some classification methods \citep{Barandela2003StrategiesProblems}, although this can be overcome by employing sampling methods, such as using only those classes with a minimum number of prototypes. The success rate of MirBot does not decrease over time although the number of classes increases, and this occurs because the number of images per class also increases. The system is, therefore, continuously improving, as more classes are detected without a decrease in the success rate. 

In summary, the main contributions of this work are: 1) an interactive workflow with which to perform object recognition and class validation; 2) a public dataset that is continuously growing with its users’ feedback, and which contains labeled images and their associated metadata; 3) an analysis of different visual descriptors (handcrafted features and neural codes from different CNN topologies) in this dataset; 4) different transfer learning methods evaluated for this task; 5) the conclusion that using normalized neural codes from the last hidden layer with kNN in the prediction stage provides a higher accuracy than the SoftMax output in all the cases evaluated; 6) approximate Nearest Neighbors techniques usually yield better efficiency and accuracy than PCA compression on neural codes; 7) metadata can be used to improve the image classification results with a two-stage architecture.


An evident future work is to further explore the usage of metadata to improve the accuracy of the visual classifier. In addition, the proposed dataset could also be analyzed using hierarchical classification methods. Although images from a root category are probably too different to benefit from hierarchical techniques, metadata could be effectively used as they may share common values in the case of certain subclasses. 

We are also planning to replace WordNet by BabelNet \cite{BabelNet} for the MirBot dictionary. BabelNet is a multilingual ontology with similar organization than WordNet but with many more concepts. Beyond having many more synsets, a multilingual dataset could be useful for example for taking a picture to some food and retrieve its name in a foreign language.

\section*{Acknowledgment} 
This work was supported by the TIMUL project (TIN2013-48152-C2-1-R) and the University Institute for Computing Research (IUII) from the University of Alicante. 


\bibliographystyle{elsarticle-num}
\bibliography{bibliography}{}

\begin{thebibliography}{10}
\expandafter\ifx\csname url\endcsname\relax
  \def\url#1{\texttt{#1}}\fi
\expandafter\ifx\csname urlprefix\endcsname\relax\def\urlprefix{URL }\fi
\expandafter\ifx\csname href\endcsname\relax
  \def\href#1#2{#2} \def\path#1{#1}\fi

\bibitem{Bock2010IImage:IPhone}
S.~Bock, S.~Newsome, Q.~Wang, W.~Zeng, X.~Lin, J.~Lu, {iImage: An image based
  information retrieval application for the iPhone}, in: 7th IEEE Consumer
  Communications and Networking Conference, CCNC, 2010, pp. 3--4.
\newblock \href {http://dx.doi.org/10.1109/CCNC.2010.5421733}
  {\path{doi:10.1109/CCNC.2010.5421733}}.

\bibitem{mobilenets}
A.~G. Howard, M.~Zhu, B.~Chen, D.~Kalenichenko, W.~Wang, T.~Weyand,
  M.~Andreetto, H.~Adam, \href{http://arxiv.org/abs/1704.04861}{{MobileNets:
  Efficient Convolutional Neural Networks for Mobile Vision Applications}},
  CoRR.
\newline\urlprefix\url{http://arxiv.org/abs/1704.04861}

\bibitem{Matusiak2013ObjectUsers}
K.~Matusiak, P.~Skulimowski, P.~Strumillo, {Object recognition in a mobile
  phone application for visually impaired users}, in: IEEE 6th International
  Conference on Human System Interactions (HSI), 2013, pp. 479--484.
\newblock \href {http://dx.doi.org/10.1109/HSI.2013.6577868}
  {\path{doi:10.1109/HSI.2013.6577868}}.

\bibitem{Fellbaum1998WordNet:Database}
C.~Fellbaum, {WordNet: An Electronic Lexical Database} (1998).
\newblock \href {http://dx.doi.org/10.1139/h11-025}
  {\path{doi:10.1139/h11-025}}.

\bibitem{Barrington2012Game-poweredLearning.}
L.~Barrington, D.~Turnbull, G.~Lanckriet, {Game-powered Machine Learning.},
  Proc. of the National Academy of Sciences of the United States of America
  109~(17) (2012) 6411--6.
\newblock \href {http://dx.doi.org/10.1073/pnas.1014748109}
  {\path{doi:10.1073/pnas.1014748109}}.

\bibitem{vonAhn2004LabelingGame}
L.~von Ahn, L.~Dabbish, {Labeling images with a computer game}, in: ACM
  Conference on Human Factors in Computing Systems, 2004, pp. 319 -- 326.
\newblock \href {http://dx.doi.org/10.1145/985692.985733}
  {\path{doi:10.1145/985692.985733}}.

\bibitem{MirBot:System}
A.~Pertusa, A.-J. Gallego, M.~Bernabeu, Mirbot: A multimodal interactive image
  retrieval system, in: {Pattern Recognition and Image Analysis}, Vol. 7887 of
  Lecture Notes in Computer Science. 6th Iberian Conference, IbPRIA, 2013, pp.
  197--204.
\newblock \href {http://dx.doi.org/10.1007/978-3-642-38628-2}
  {\path{doi:10.1007/978-3-642-38628-2}}.

\bibitem{Torralba200880Recognition.}
A.~Torralba, R.~Fergus, W.~T. Freeman, {80 Million Tiny Images: a Large Data
  Set for Nonparametric Object and Scene Recognition.}, IEEE Trans. on Pattern
  Analysis and Machine Intelligence (PAMI) 30~(11) (2008) 1958--1970.
\newblock \href {http://dx.doi.org/10.1109/TPAMI.2008.128}
  {\path{doi:10.1109/TPAMI.2008.128}}.

\bibitem{Khosla2012UndoingBias}
A.~Khosla, T.~Zhou, T.~Malisiewicz, A.~A. Efros, A.~Torralba, {Undoing the
  damage of dataset bias}, Lecture Notes in Computer Science 7572 LNCS~(part 1)
  (2012) 158--171.
\newblock \href {http://dx.doi.org/10.1007/978-3-642-33718-5_12}
  {\path{doi:10.1007/978-3-642-33718-5_12}}.

\bibitem{Deng2009ImageNet:Database}
J.~Deng, W.~Dong, R.~Socher, L.-J. Li, K.~Li, L.~Fei-Fei, {ImageNet: A
  large-scale hierarchical image database}, in: IEEE Conference on Computer
  Vision and Pattern Recognition (CVPR), 2009, pp. 2--9.
\newblock \href {http://dx.doi.org/10.1109/CVPR.2009.5206848}
  {\path{doi:10.1109/CVPR.2009.5206848}}.

\bibitem{LeCun2015DeepLearning}
Y.~LeCun, Y.~Bengio, G.~Hinton, Deep learning, Nature 521~(7553) (2015)
  436--444.
\newblock \href {http://dx.doi.org/10.1038/nature14539}
  {\path{doi:10.1038/nature14539}}.

\bibitem{Lew2006Content-basedChallenges}
M.~Lew, N.~Sebe, C.~Djeraba, R.~Jain, {Content-based multimedia information
  retrieval: State of the art and challenges}, ACM Trans. on Multimedia
  Computing, Communications, and Applications 2~(1) (2006) 1--19.
\newblock \href {http://dx.doi.org/10.1145/1126004.1126005}
  {\path{doi:10.1145/1126004.1126005}}.

\bibitem{Mikolajczyk2004ScaleDetectors}
K.~Mikolajczyk, C.~Schmid, {Scale {\&} affine invariant interest point
  detectors}, International Journal of Computer Vision 60~(1) (2004) 63--86.
\newblock \href {http://dx.doi.org/10.1023/B:VISI.0000027790.02288.f2}
  {\path{doi:10.1023/B:VISI.0000027790.02288.f2}}.

\bibitem{Lowe2004DistinctiveKeypoints}
D.~G. Lowe, {Distinctive image features from scale-invariant keypoints},
  International Journal of Computer Vision 60~(2) (2004) 91--110.
\newblock \href {http://dx.doi.org/10.1023/B:VISI.0000029664.99615.94}
  {\path{doi:10.1023/B:VISI.0000029664.99615.94}}.

\bibitem{Bay2008Speeded-UpSURF}
H.~Bay, A.~Ess, T.~Tuytelaars, L.~{Van Gool}, {Speeded-Up Robust Features
  (SURF)}, Computer Vision and Image Understanding 110~(3) (2008) 346--359.
\newblock \href {http://dx.doi.org/10.1016/j.cviu.2007.09.014}
  {\path{doi:10.1016/j.cviu.2007.09.014}}.

\bibitem{vandeSande2010EvaluatingRecognition}
K.~E.~A. van~de Sande, T.~Gevers, C.~G.~M. Snoek, {Evaluating Color Descriptors
  for Object and Scene Recognition}, IEEE Trans. on Pattern Analysis and
  Machine Intelligence (PAMI) 32~(9) (2010) 1582--1596.
\newblock \href {http://dx.doi.org/10.1109/TPAMI.2009.154}
  {\path{doi:10.1109/TPAMI.2009.154}}.

\bibitem{Jeong2001Histogram-BasedRetrieval}
S.~Jeong,
  \href{https://pdfs.semanticscholar.org/e884/2a22aa486fd273606fcd5f8090619bbb468c.pdf}{{Histogram-Based
  Color Image Retrieval}}, Tech. rep., Psych221/EE362 Project Report, Stanford
  (2001).
\newline\urlprefix\url{https://pdfs.semanticscholar.org/e884/2a22aa486fd273606fcd5f8090619bbb468c.pdf}

\bibitem{Fernando2012DiscriminativeClassification}
B.~Fernando, E.~Fromont, D.~Muselet, M.~Sebban, {Discriminative feature fusion
  for image classification}, in: IEEE Conf. on Computer Vision and Pattern
  Recognition (CVPR), 2012, pp. 3434--3441.
\newblock \href {http://dx.doi.org/10.1109/cvpr.2012.6248084}
  {\path{doi:10.1109/cvpr.2012.6248084}}.

\bibitem{Banerji2013NewClassification}
S.~Banerji, A.~Sinha, C.~Liu, {New image descriptors based on color, texture,
  shape, and wavelets for object and scene image classification},
  Neurocomputing 117 (2013) 173--185.
\newblock \href {http://dx.doi.org/10.1016/j.neucom.2013.02.014}
  {\path{doi:10.1016/j.neucom.2013.02.014}}.

\bibitem{Thomee2010TOP-SURF:Toolkit}
B.~Thomee, E.~M. Bakker, M.~S. Lew, {TOP-SURF: A Visual Words Toolkit}, ACM
  International Conference on Multimedia\href
  {http://dx.doi.org/10.1145/1873951.1874250}
  {\path{doi:10.1145/1873951.1874250}}.

\bibitem{Philbin2007ObjectMatching}
J.~Philbin, O.~Chum, M.~Isard, J.~Sivic, A.~Zisserman, {Object retrieval with
  large vocabularies and fast spatial matching}, in: {IEEE Conf. on Computer
  Vision and Pattern Recognition (CVPR)}, 2007.
\newblock \href {http://dx.doi.org/10.1109/CVPR.2007.383172}
  {\path{doi:10.1109/CVPR.2007.383172}}.

\bibitem{Salton1983IntroductionRetrieval.}
G.~Salton, M.~J. McGill, {Introduction to modern information retrieval},
  McGraw-Hill, Inc, New York, NY, USA, 1986.
\newblock \href {http://dx.doi.org/10.1108/01435121111132365}
  {\path{doi:10.1108/01435121111132365}}.

\bibitem{Yosinski2014How}
J.~Yosinski, J.~Clune, Y.~Bengio, H.~Lipson, {How transferable are features in
  deep neural networks?}, in: Proc. of Neural Information Processing Systems
  (NIPS), 2014.
\newblock \href {http://arxiv.org/abs/1411.1792} {\path{arXiv:1411.1792}}.

\bibitem{Krizhevsky2012ImageNetNetworks}
A.~Krizhevsky, I.~Sutskever, G.~E. Hinton, {ImageNet Classification with Deep
  Convolutional Neural Networks}, in: Proc. Neural Information and Processing
  Systems (NIPS), 2012.
\newblock \href {http://arxiv.org/abs/1102.0183} {\path{arXiv:1102.0183}}.

\bibitem{Russakovsky2014ImageNetChallenge}
O.~Russakovsky, J.~Deng, H.~Su, J.~Krause, S.~Satheesh, S.~Ma, Z.~Huang,
  A.~Karpathy, A.~Khosla, M.~Bernstein, A.~C. Berg, L.~Fei-Fei, {ImageNet Large
  Scale Visual Recognition Challenge}, {International Journal of Computer
  Vision (IJCV)}\href {http://dx.doi.org/10.1007/s11263-015-0816-y}
  {\path{doi:10.1007/s11263-015-0816-y}}.

\bibitem{Szegedy2014GoingConvolutions}
C.~Szegedy, W.~Liu, Y.~Jia, P.~Sermanet, S.~Reed, D.~Anguelov, D.~Erhan,
  V.~Vanhoucke, A.~Rabinovich, {Going Deeper with Convolutions}, in: IEEE Conf.
  on Computer Vision and Pattern Recognition (CVPR), 2015.
\newblock \href {http://arxiv.org/abs/1409.4842} {\path{arXiv:1409.4842}}.

\bibitem{Simonyan14}
K.~Simonyan, A.~Zisserman, Very deep convolutional networks for large-scale
  image recognition, 2014.
\newblock \href {http://arxiv.org/abs/1409.1556} {\path{arXiv:1409.1556}}.

\bibitem{Ioffe15:ArXiv}
S.~Ioffe, C.~Szegedy, Batch normalization: Accelerating deep network training
  by reducing internal covariate shift, 2015.
\newblock \href {http://arxiv.org/abs/1502.03167} {\path{arXiv:1502.03167}}.

\bibitem{inceptionv3}
C.~Szegedy, V.~Vanhoucke, S.~Ioffe, J.~Shlens, Z.~Wojna,
  \href{http://arxiv.org/abs/1512.00567}{Rethinking the inception architecture
  for computer vision}, CoRR abs/1512.00567.
\newline\urlprefix\url{http://arxiv.org/abs/1512.00567}

\bibitem{resnet}
K.~He, X.~Zhang, S.~Ren, J.~Sun, \href{http://arxiv.org/abs/1512.03385}{Deep
  residual learning for image recognition}, CoRR abs/1512.03385.
\newline\urlprefix\url{http://arxiv.org/abs/1512.03385}

\bibitem{xception}
F.~Chollet, \href{http://arxiv.org/abs/1610.02357}{Xception: Deep learning with
  depthwise separable convolutions}, CoRR abs/1610.02357.
\newline\urlprefix\url{http://arxiv.org/abs/1610.02357}

\bibitem{Jia2014CaffeEmbedding}
Y.~Jia, E.~Shelhamer, J.~Donahue, S.~Karayev, J.~Long, R.~Girshick,
  S.~Guadarrama, T.~Darrell, {Caffe : Convolutional Architecture for Fast
  Feature Embedding}, in: ACM Conference on Multimedia, 2014, pp. 675--678.
\newblock \href {http://dx.doi.org/10.1145/2647868.2654889}
  {\path{doi:10.1145/2647868.2654889}}.

\bibitem{MXNet}
T.~Chen, M.~Li, N.~Wang, M.~a. Wang, Mxnet: A flexible and efficient machine
  learning library for heterogeneous distributed systems, 2016.
\newblock \href {http://arxiv.org/abs/1512.01274} {\path{arXiv:1512.01274}}.

\bibitem{keras}
F.~Chollet, Keras, \url{https://github.com/fchollet/keras} (2015).

\bibitem{He09}
H.~He, E.~Garcia, Learning from imbalanced data, IEEE Trans. Knowl. Data Eng.
  21 (2009) 1263--1284.
\newblock \href {http://dx.doi.org/10.1109/TKDE.2008.239}
  {\path{doi:10.1109/TKDE.2008.239}}.

\bibitem{annoy}
E.~Bernhardsson, \href{https://github.com/spotify/annoy}{Annoy: Approximate
  nearest neighbors in c++/python optimized for memory usage and loading/saving
  to disk} (2016).
\newline\urlprefix\url{https://github.com/spotify/annoy}

\bibitem{Lin1991DivergenceEntropy}
J.~Lin, {Divergence measures based on the Shannon entropy}, IEEE Trans. on
  Information Theory 37~(1) (1991) 145--151.
\newblock \href {http://dx.doi.org/10.1109/18.61115}
  {\path{doi:10.1109/18.61115}}.

\bibitem{BabenkoSCL14}
A.~Babenko, A.~Slesarev, A.~Chigorin, V.~S. Lempitsky,
  \href{http://arxiv.org/abs/1404.1777}{Neural codes for image retrieval}, CoRR
  abs/1404.1777.
\newblock \href {http://arxiv.org/abs/1404.1777} {\path{arXiv:1404.1777}}.
\newline\urlprefix\url{http://arxiv.org/abs/1404.1777}

\bibitem{Sparse}
A.~Parashar, M.~Rhu, A.~Mukkara, A.~Puglielli, R.~Venkatesan, B.~Khailany,
  J.~S. Emer, S.~W. Keckler, W.~J. Dally,
  \href{http://arxiv.org/abs/1708.04485}{{SCNN:} an accelerator for
  compressed-sparse convolutional neural networks}, CoRR abs/1708.04485.
\newblock \href {http://arxiv.org/abs/1708.04485} {\path{arXiv:1708.04485}}.
\newline\urlprefix\url{http://arxiv.org/abs/1708.04485}

\bibitem{weka}
M.~Hall, E.~Frank, G.~Holmes, B.~Pfahringer, P.~Reutemann, I.~H. Witten,
  \href{http://doi.acm.org/10.1145/1656274.1656278}{The weka data mining
  software: An update}, SIGKDD Explor. Newsl. 11~(1) (2009) 10--18.
\newblock \href {http://dx.doi.org/10.1145/1656274.1656278}
  {\path{doi:10.1145/1656274.1656278}}.
\newline\urlprefix\url{http://doi.acm.org/10.1145/1656274.1656278}

\bibitem{feat_codes}
\href{http://www.geonames.org/export/codes.html}{{Geonames feature codes}}.
\newline\urlprefix\url{http://www.geonames.org/export/codes.html}

\bibitem{Mao2014ExplainNetworks}
J.~Mao, W.~Xu, Y.~Yang, J.~Wang, A.~L. Yuille, {Explain Images with Multimodal
  Recurrent Neural Networks}, in: NIPS Deep Learning Workshop, 2014, pp. 1--9.
\newblock \href {http://arxiv.org/abs/1410.1090v1} {\path{arXiv:1410.1090v1}}.

\bibitem{Bellet2012SimilarityClassification}
A.~Bellet, A.~Habrard, M.~Sebban, {Similarity Learning for Provably Accurate
  Sparse Linear Classification}, in: 29th International Conference on Machine
  Learning, 2012.
\newblock \href {http://arxiv.org/abs/1206.6476} {\path{arXiv:1206.6476}}.

\bibitem{Barandela2003StrategiesProblems}
R.~Barandela, J.-S. S{\'{a}}nchez, V.~Garc\'{i}a, E.~Rangel, {Strategies for
  learning in class imbalance problems}, Pattern Recognition 36~(3) (2003)
  849--851.
\newblock \href {http://dx.doi.org/10.1016/S0031-3203(02)00257-1}
  {\path{doi:10.1016/S0031-3203(02)00257-1}}.

\bibitem{BabelNet}
R.~Navigli, S.~P. Ponzetto, {B}abel{N}et: {T}he automatic construction,
  evaluation and application of a wide-coverage multilingual semantic network,
  Artificial Intelligence 193 (2012) 217--250.

\end{thebibliography}

\end{document}